%% file: main.tex
\documentclass[]{spie}  

\usepackage{amsmath,amsfonts,amssymb}
\usepackage{graphicx}
\usepackage{pifont}%
\usepackage{adjustbox}
\usepackage{multirow}

\usepackage{float} 
\usepackage{xcolor}
\usepackage[colorlinks=true, allcolors=blue]{hyperref}

\title{GTNet: Guided Transformer Network for Detecting Human-Object Interactions}

\author[a]{A S M Iftekhar}
\author[a]{Satish Kumar}
\author[a]{R. Austin  McEver}
\author[b]{Suya You}
\author[a]{B.S. Manjunath}
\affil[a]{Department of Electrical \& Computer Engineering, University of California, Santa Barbara}
\affil[b]{US Army Research Laboratory}

\authorinfo{Further author information: (Send correspondence to A. I.)\\A.I.: E-mail: iftekhar@ucsb.edu, \\  
S.K.: E-mail: satishkumar@ucsb.edu, \\
A.M.: E-mail: mcever@ucsb.edu, \\
S.Y.: E-mail: suya.you.civ@army.mil, \\
B.M.: E-mail: manj@ucsb.edu.
}

\begin{document} 
\maketitle
\input{content/1_abstract_pub}


\input{content/2_introduction}
\input{content/3_related_works}
\input{content/4_methodology_new}
\input{content/5_experiments}

\input{content/6_results}

\input{content/7_conclusions}
\input{content/Acknowledgement}

\bibliography{report} 

\bibliographystyle{spiebib} 
\include{content/supp}
\end{document}

%% file: content/1_abstract_pub.tex
\begin{abstract}
The human-object interaction (HOI) detection task refers to localizing humans, localizing objects, and predicting the interactions between each human-object pair. HOI is considered one of the fundamental steps in truly understanding complex visual scenes. For detecting HOI, it is important to utilize relative spatial configurations and object semantics to find salient spatial regions of images that highlight the interactions between human object pairs. This issue is addressed by the novel self-attention based guided transformer network, GTNet. GTNet encodes this spatial contextual information in human and object visual features via self-attention while achieving state of the art results on both the V-COCO~\cite{gupta2015visual} and HICO-DET~\cite{chao2018learning} datasets. Code is available online\footnote{\href{https://github.com/UCSB-VRL/GTNet}{https://github.com/UCSB-VRL/GTNet}}.  
 \end{abstract}

%% file: content/2_introduction.tex
\section{INTRODUCTION}

Understanding human activity from images and videos is one of the main goals of computer vision based systems. One of the first steps in this process is the successful detection of human-object interactions (HOIs) in images, which consists of detecting the interactions between a human and an object while localizing their respective locations. In recent times, HOI has received much attention in the computer vision community due to its importance in tasks like action recognition in videos~\cite{feichtenhofer2019slowfast,li2019collaborative,ulutan2020actor}, scene understanding~\cite{ren2019deep,chen2019towards,behley2019semantickitti}, and visual question answering~\cite{chen2020counterfactual,cadene2019murel,li2019relation}. 
\begin{figure}[t]
\centering
\includegraphics[width=0.8\linewidth]{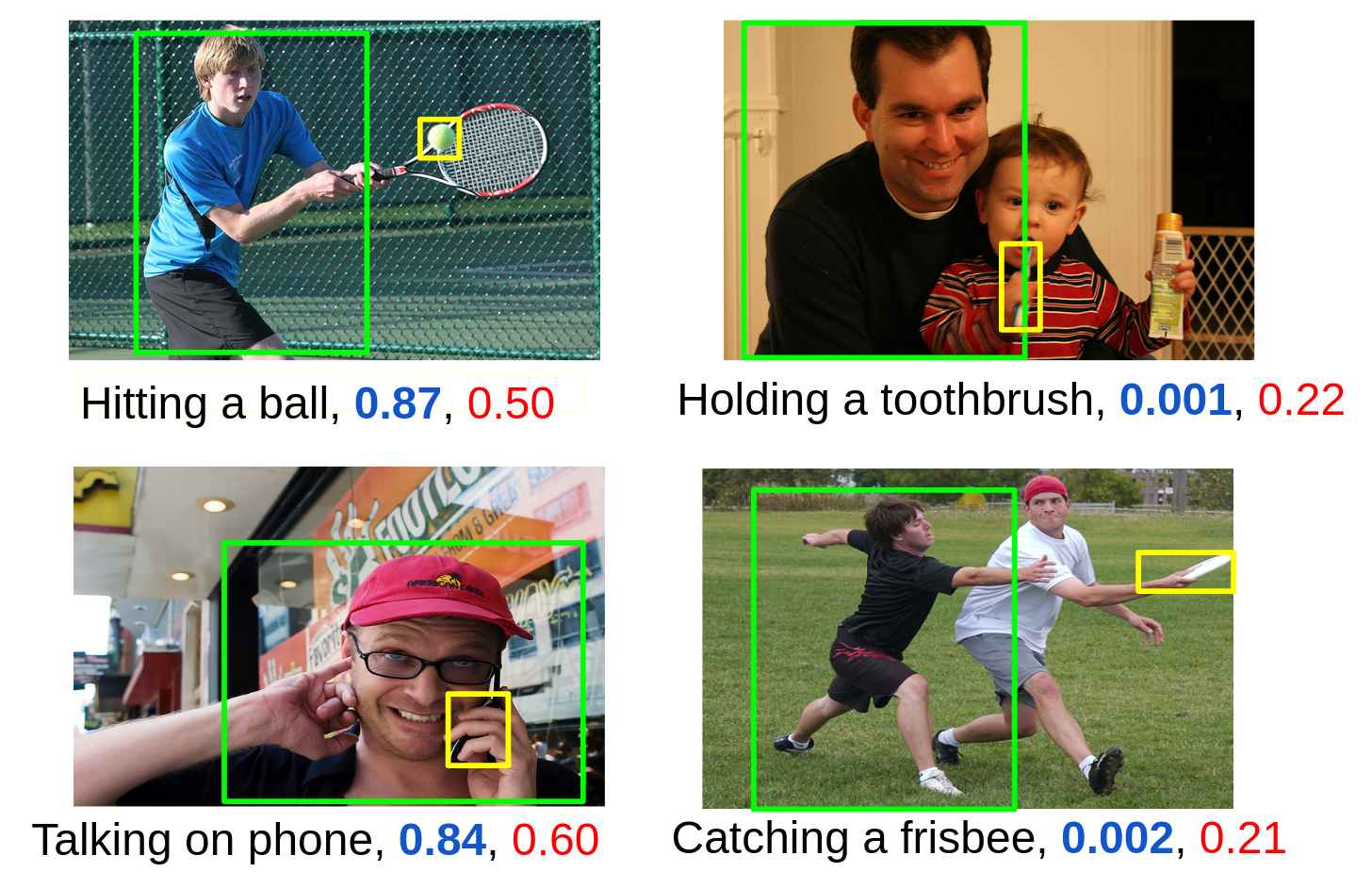}
   \caption{GTNet's performance with and without its guidance mechanism. Blue indicates GTNet's predictions with the guidance mechanism; red indicates without. Green bounding boxes indicate the human under consideration; yellow boxes indicate the object. Left column: Simple Scenarios (higher confidence score is better). Right column: Potentially Confusing Scenarios (lower confidence score is better). With the guidance mechanism, it is easier find salient spatial context for detecting different types of HOIs. }
\label{fig:short}
\end{figure}
Current research has integrated many different techniques ranging from attention mechanisms~\cite{ulutan2020vsgnet,gao2018ican} to graph convolutional networks~\cite{qi2018learning,gao2020drg} to detect HOIs. In basic detection frameworks, human and object features are usually extracted with an object detection network. Then, interactions are predicted from the extracted features. To strengthen these features, one could use additional features such as relative spatial configurations~\cite{ulutan2020vsgnet, gao2018ican}, human pose estimation~\cite{wan2019pose}, or more accurate segmentation masks~\cite{liuamplifying}. 

However, to fully utilize the power of these additional features, the local spatial context needs to be leveraged more effectively. In computer vision tasks, context often refers directly to the background and surroundings of the objects or people of interest. In the following, we use "context" to refer to spatial regions localizing the human-object interactions. Few current works~\cite{wang2019deep,gao2018ican, gao2020drg, wang2020contextual} have tried to find relevant spatial contexts by having separate attention mechanisms for humans and objects but do not leverage any additional features in the attention framework. 
To this end, GTNet proposes a unique way to find spatial contextual information for detecting HOIs by leveraging spatial configurations (i.e. relative spatial layout between a human and an object) and object semantics (i.e. category of objects represented by word embeddings). This method has proven to be very effective, as can be seen in Figure~\ref{fig:short}, where our proposed Guidance Module improves performance over our model without the guidance mechanism. 

We take inspiration from Natural Language Processing (NLP) where the self attention~\cite{xu2015show} based Transformer architecture~\cite{vaswani2017attention} has shown significant success in finding contextual information. 
To adapt the Transformer architecture for detecting HOIs, GTNet concatenates pairwise human and object features and uses them as queries to the attention mechanism. However, detection of HOIs often depend upon relative spatial configuration~\cite{ulutan2020vsgnet} and the type of the objects\cite{liu2020consnet}. Therefore, we combine relative spatial configurations with object semantics to guide the queries throughout the network. With the help of our proposed guided attention, we successfully encode the spatial contextual information in these queries.


 
The proposed GTNet architecture can be seen in Figure~\ref{fig:model_architecture}. From the input image and the bounding boxes of the humans and the objects present in it, our baseline module (Section~\ref{baseline}) extracts visual features. These types of visual features from a backbone network are being used across several domains~\cite{kumar2021stressnet, wang2018comparative, kumar2020deep} to capture useful information. We guide these visual features by pertinent spatial configurations and object semantics in our Guidance Module (Section~\ref{guide}). On top of these guided visual features, we develop the TX module, which enriches the visual features with relevant contextual information using attention (Section~\ref{TX_Mod}). Finally, we propose an early fusion strategy to make our final predictions (Section~\ref{fuse}).
We evaluated our network's performance on V-COCO~\cite{gupta2015visual} and HICO-DET~\cite{chao2018learning} and achieve state of the art results on both of the datasets. 
Our contributions can be summarized as follows:
 \begin{itemize}
  \item We leverage pairwise spatial contextual information via a novel end to end guided self attention network for detecting HOIs. See Section~\ref{TX_Mod}.  
  \item We design a guidance mechanism that combines relative spatial configurations and object semantics to guide our attention mechanism. See section~\ref{guide}. 
  \item GTNet achieves state of the art results for the HOI detection task on both V-COCO and HICO-DET datasets. See section~\ref{section:results}. 
  \end{itemize}



%% file: content/3_related_works.tex
\section{Related Works}
\paragraph{\textbf{Human-Object Interaction:}} With the introduction of benchmark datasets like V-COCO~\cite{gupta2015visual} and  HICO-DET~\cite{chao2018learning}, there is a plethora of works detecting human-object interactions~\cite{fang2020dirv,gao2018ican,li2019transferable,liuamplifying, wan2019pose, wang2019deep,gao2020drg,ulutan2020vsgnet, qi2018learning, zhong2020polysemy, kim2020detecting, liu2020consnet, hou2020visual, wang2020contextual, IDN}. Earlier works~\cite{gkioxari2018detecting} in this area focus on the visual features of humans and objects. Many subsequent works~\cite{wang2019deep,gao2018ican, gao2020drg} try to find spatial context for interactions on top of these visual features. Gao et al.~\cite{gao2018ican} present a self-attention mechanism around individual humans and objects. T. Wang et al.~\cite{wang2019deep} leverage this attention mechanism with a squeeze and excitation block~\cite{hu2018squeeze} from object detection. Recently, graph-based architectures where humans and objects are considered nodes attempt to understand spatial context~\cite{gao2020drg,wang2020contextual} for the structural relations. ConsNet~\cite{liu2020consnet} has leveraged word embeddings of the objects in this graph structure. Additionally, Hou et al.~\cite{hou2021affordance, hou2021detecting} have utilized object affordance to detect HOIs. A few recent works ~\cite{tamura2021qpic, chen2021reformulating, kim2021hotr, zou2021end} have developed a one stage pipeline to detect HOIs rather than the two-stage (object detection + HOI detection) approach. We only compare our method with two stage HOI detection networks.
However, none of the existing attention-based works try to utilize additional features like relative spatial configurations or object semantics to find richer spatial contextual features in the attention framework. In this aspect, GTNet proposes a pairwise attention network with a guidance mechanism for detecting HOIs by encoding spatial contextual information. We leverage spatial configurations and object semantic information to guide our attention network. 

Many of the current works also use different additional features~\cite{chao2018learning,ulutan2020vsgnet,li2019transferable,liuamplifying,wan2019pose,li2020detailed} ranging from pose information of humans to 3D representations of 2D images for detecting HOIs. Recently, VSGNet~\cite{ulutan2020vsgnet} utilizes relative spatial configurations to refine visual features. Inspired by this approach, we use spatial configurations to guide our self-attention mechanism. Further, we combine object semantics~\cite{peyre2019detecting,liu2020consnet} with spatial configurations~\cite{chao2018learning} for our guidance system. Our empirical results show that this combination performs better as a guidance mechanism.
\vspace{-0.5cm}
\paragraph{\textbf{Transformer Network:}} 
Recently, Transformer~\cite{vaswani2017attention} based networks have achieved the state of the art performances in different vision tasks~\cite{carion2020end,dosovitskiy2020image, tamura2021qpic}. For detecting HOIs, one of the first attempts to use transformer like self-attention was~\cite{girdhar2017attentional}. Following that work, Girdher et al.~\cite{girdhar2019video} has developed the Actor Transformer network for detecting human activities in videos. Few recent works~\cite{tamura2021qpic, chen2021reformulating, kim2021hotr, zou2021end} have developed one stage Transformer networks for detecting HOIs. However, all these works rely only on self-attention mechanism to find salient context in the Transformer architecture. In contrast, we guide our joint feature representations of each human-object pair with spatial configurations and object semantics to find salient spatial context. Our state of the art performance over standard datasets validates our design choices for the self-attention network.  



%% file: content/4_methodology_new.tex
\begin{figure*}[t]
\begin{center}
\includegraphics[width=1.0\linewidth]{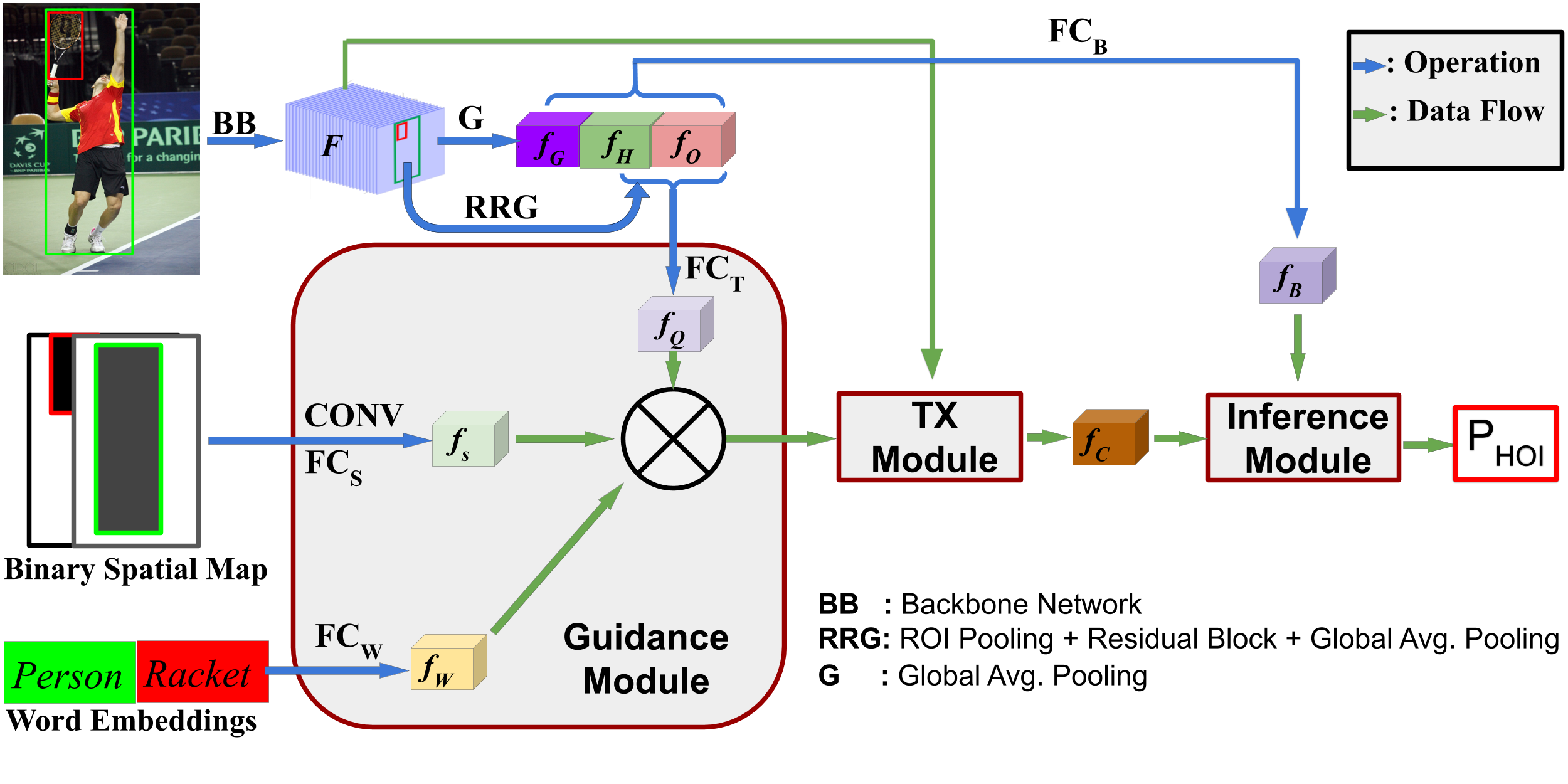}
\end{center}
\vspace{-0.4cm}
   \caption{Model Overview. We extract human and object feature vectors from the input feature map, $\mathbf{F}$ via two RRG operations. For clarity we do not explicitly show two separate RRG operations. Human and object features are used to generate a query vector,  $\mathbf{f}_{Q}$. Before feeding $\mathbf{f}_{Q}$ to the TX Module, we guide it via an element wise product with spatial ($\mathbf{f}_{S}$) and semantic ($\mathbf{f}_{W}$) guidance feature vectors. Inside the TX Module, contextual information is encoded to the guided query vector to generate a context-aware updated query vector $\textbf{f}_{C}$. Finally, we make HOI predictions ($\mathbf{P}_{HOI}$) from the updated query and the baseline feature vectors in the Inference Module. Details of TX Module in Figure~\ref{fig:tx_unit}.
   }
\label{fig:model_architecture}
\vspace{-0.4cm}
\end{figure*}
\section{Technical Approach}
In this section, we introduce our proposed GTNet architecture for Human-Object interaction (HOI) detection. Given an input image, the task is to generate bounding boxes for all humans and objects while detecting the interactions among them (e.g. a person hitting a ball). Each human-object pair can have multiple interactions. We use a pre-trained object detector to detect humans and objects in the images. 

GTNet takes image features $\textbf{F}$  as input. For extracting features, we use standard feature extractor networks (e.g., resnet~\cite{he2016deep} and  efficientnet~\cite{tan2019efficientnet}). Consider a single image of size $c\times h\times w$ where $c,h,$ and $w$ are the number of channels, height and width of the image. For a given image, $\textbf{F}$ has a dimension of $C\times H\times W$ in the feature space. The relationship between $C,H,W$ and $c,h, w$ depends on the backbone network. 

Along with image features, for each human and object our network takes bounding boxes $b_h$ and $b_o$ as inputs when there are $M\geq1$ humans and $N\geq1$ objects present in the image. As mentioned earlier, we use a pre-trained object detector to generate those bounding boxes. Our network will predict an interaction probability vector $\mathbf{p}_{HOI}$ for each human-object pair. In the next sections, we will describe different components of our network.

\subsection{Baseline Module} \label{baseline}
Our baseline module extracts human and object feature vectors $\mathbf{f}_{H}$, and $\mathbf{f}_{O}$ from the input feature map $\mathbf{F}$. Following previous works~\cite{gao2018ican,ulutan2020vsgnet} we use region of interest pooling followed by a residual block and average pooling to extract these feature vectors. Moreover, we use the overall feature map $\mathbf{F}$ to get generalized context information by extracting feature vector $\mathbf{f}_{G}$ with average pooling. Recent works like VSGNet~\cite{ulutan2020vsgnet} use this feature vector as a representation of context, we argue that it is not rich enough to account for interaction specific contexts. This is why we propose the guided attention mechanism to encode task specific spatial contextual information.

To get a joint representation of these feature vectors, we concatenate and project them in the same space using a fully connected (FC) layer.
\begin{equation}
    \mathbf{f}_{B} = \mathbf{FC}_{B}(\mathbf{f}_{H}  \mathbin\Vert \mathbf{f}_{O}  \mathbin\Vert \mathbf{f}_{G})
\end{equation}
Here, $ \mathbin\Vert$ represents concatenation. 
This naive feature vector $\mathbf{f}_{B}$ is not adequate to detect all fine-grained HOIs. In the next sections, we describe how to refine and couple visual features with human-object pairwise spatial contextual information.

\subsection{Guidance Module} \label{guide}
Vaswani et al.~\cite{vaswani2017attention} propose the Transformer network for processing sequential data in natural language processing (NLP). This network uses self attention~\cite{xu2015show} to find contextual dependence in sequential data. The original Transformer network introduces the concept of queries, keys, and values, which we adopt to the context of HOI detection. In this section, we explain the idea of queries and the mechanism to guide it. In the next section, we explain our attention mechanism.

Queries are defined as the pairwise joint representation of human and object feature vectors ($\mathbf{f}_{H}$, $\mathbf{f}_{O}$). We get the pairwise representation by concatenation and projection.
\begin{equation}
    \mathbf{f}_{Q} = \mathbf{FC}_{T}(\mathbf{f}_{H}  \mathbin\Vert \mathbf{f}_{O})
\end{equation}
Here, $\mathbf{f}_{Q}$ is the query vector, and $ \mathbin\Vert$ represents concatenation operation. For a single human-object pair query vector has a length of $D$. This query vector will be used to find relevant spatial context in the feature map.

As mentioned in many previous works~\cite{vaswani2017attention,girdhar2019video}, Transformer-like architectures do not perform well without positional information embedded in queries. For action recognition, Girdhar et al.~\cite{girdhar2019video} embedded bounding box sizes and locations in the queries. HOI detection is a more subtle task as, along with the size of the humans and the objects, relative configurations among them are also important. To this end, we propose our guidance module that uses relative spatial configurations and semantic representations of humans and objects to guide the attention mechanism. 

\noindent{\textbf{Spatial Guidance:}}
Relative spatial configurations have proven to be very useful~\cite{chao2018learning,ulutan2020vsgnet,li2019transferable,gao2018ican} in detecting HOIs. We use a two-channel binary map with a dimension of $2\times s\times s$ ($s$ is chosen as 64, see Table~\ref{tab:new_bin_map}) to encode relative spatial configurations. For a human-object pair, the first channel is 1 in the location of the human-bounding box, whereas the second channel contains 1 in the location of the object bounding box. Everywhere else is zero in the binary map. Following~\cite{chao2018learning}, we use two convolutional layers along with average pooling and linear projection($\mathbf{FC}_{S}$) to get a feature vector $\mathbf{f}_{S}$, representing the relative spatial configurations. We utilize $\mathbf{f}_{S}$ to guide the TX Module.

\noindent{\textbf{Semantic Guidance:}} Although $\mathbf{f}_{S}$ is a strong cue to detect HOIs, the information it conveys can be confusing without proper knowledge of the object. That is why we combine object semantics with spatial configurations in our guidance mechanism. Instead of just using a look-up table for identifying each object, we use word embeddings (vector representation of words) from the publicly available Glove~\cite{pennington2014glove} model. 
For our specific case, every detected object by the object detector is represented with a vector. For example, phone and football are presented by two different vectors. We express humans with one fixed vector.
After concatenation of these human and object word embedding vectors together we get a combined feature vector $\mathbf{f}_{W}$ with linear projection ($\mathbf{FC}_{W}$). Along with $\mathbf{f}_{S}$, we utilize $\mathbf{f}_{W}$ in the guiding mechanism. 

\noindent{\textbf{Guidance Mechanism:}} \label{guide_mech}The information present in $\mathbf{f}_{S}$ and $\mathbf{f}_{W}$ is used to guide the queries sent to the TX Module. Guiding here refers to encoding relative spatial configurations and object semantics to the queries before feeding them to the TX Module. This can be achieved either by concatenation or by taking the element-wise product among the spatial, semantic ,and the query vectors. We use element-wise product as it is proven to be more effective guidance mechanism (See Table ~\ref{tab:ablation_study}) than concatenation. 
\begin{align}
    \mathbf{f}_{GQ} = {\mathbf{f}_{Q}\circ \mathbf{f}_{S}\circ \mathbf{f}_{W}} 
    \label{eq:HADA}
\end{align}
Here, $\circ$ represents element wise dot product. We feed the guided query vector $\mathbf{f}_{GQ}$ to the TX Module. 

\subsection{TX Module}\label{TX_Mod}
 
   \begin{figure}[t]
\begin{center}
\includegraphics[width=0.6\linewidth]
{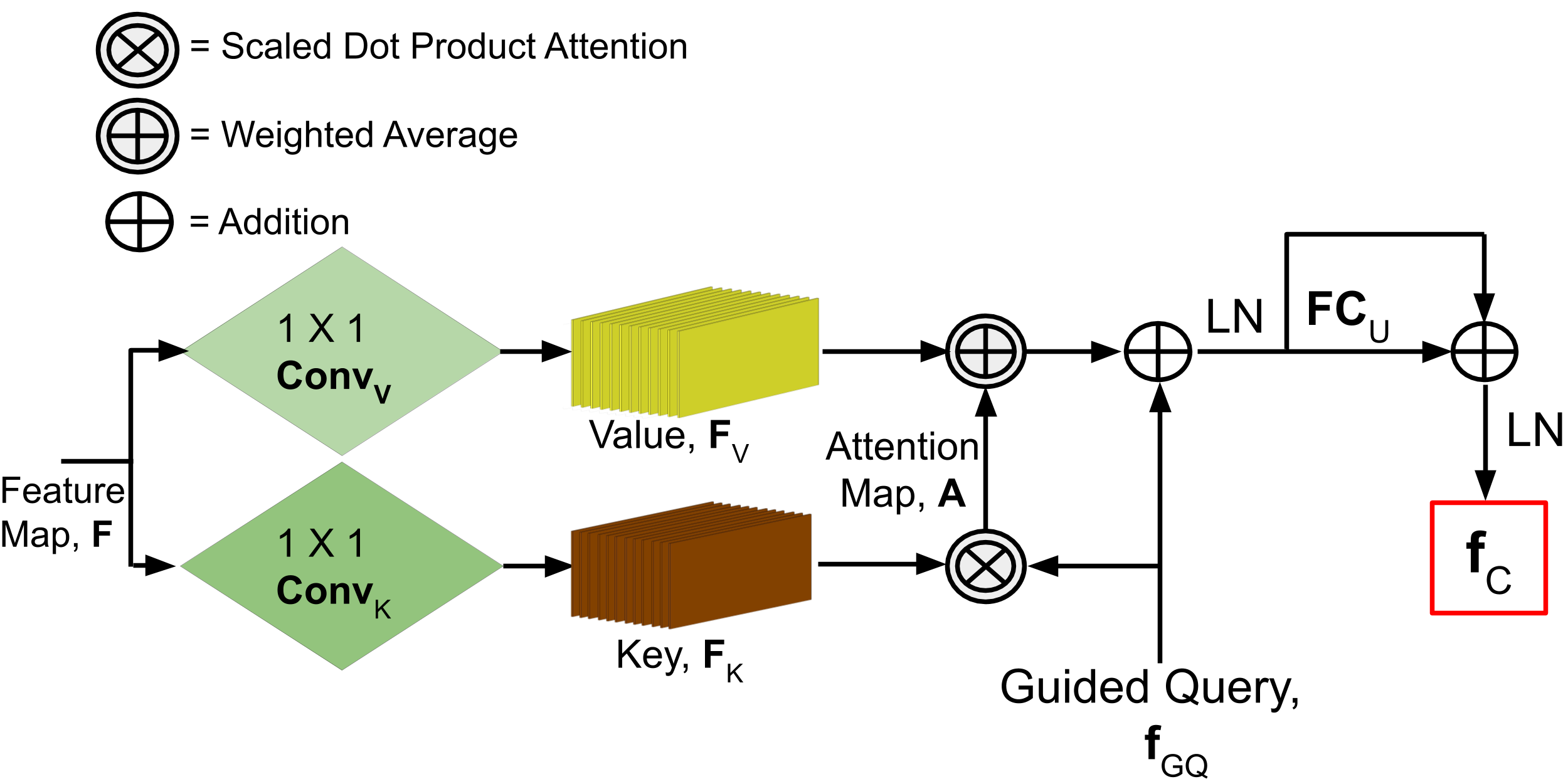}
\end{center}
\vspace{-0.4cm}
\caption{TX Module. From the input feature map, we generate key and value by $1\times 1$ convolutions.  With scaled dot-product attention, we produce an attention map for a particular human-object pair from the query and the key. This attention map is used to weigh the value to derive the contextually rich feature vector $\textbf{f}_{C}$. }
\vspace{-0.4cm}
\label{fig:tx_unit}
\end{figure}
The TX module is our adaptation of the original Transformer architecture. Figure~\ref{fig:tx_unit} shows the details of TX Module. This module encodes spatial contextual information to the guided query vectors via attention. We leverage the concept of keys and values from the original Transformer architecture in this mechanism.   

Keys ($\mathbf{F}_{K}$) and values ($\mathbf{F}_{V}$) are generated from the input feature map ($\mathbf{F}$) by two separate $1\times1$ convolutions. 
\begin{flalign}
    &\mathbf{F}_{K} = \mathbf{conv}_{K}(\mathbf{F}) \\
    &\mathbf{F}_{V} = \mathbf{conv}_{V}(\mathbf{F})
\end{flalign}

Here, $\mathbf{conv}_{K}$ and $\mathbf{conv}_{V}$ are two different $1\times1$ convolution operations. $\mathbf{F}_{K}$ and $\mathbf{F}_{V}$ can be thought of as two different representations of the input feature map. For a single image, both of them have the dimension of $D\times H\times W$. 

The guided query vectors are used to search keys, $\mathbf{F}_{K}$ to find pairwise contextual information. Imagine a person is talking on phone in an image. Our guided query vector representing the person and the phone pair finds the important region (i.e. close to the ear) in $\mathbf{F}_{K}$ to detect the interaction: talking on phone. 

This search process is done by scaled dot-product attention (equ.~\ref{dot}). In each spatial location of $\mathbf{F}_{K}$, we take a channel-wise scaled dot-product with the guided query vector. Softmax is applied to present the important context for a particular guided query in probabilities.  


\begin{align}
    \mathbf{A} = \textbf{Softmax}(\frac{\mathbf{f}_{GQ}\mathbf{F}_{K}^{T}}{\sqrt{\mathbf{D}}}) 
    \label{dot}
\end{align}
For a particular guided query, $\mathbf{A}$ is the attention map where each element signifies the probability of that spatial location to be significant for detecting interactions. $\mathbf{A}$ has a dimension of $H\times W$ for each guided query vector. We use the attention map $\mathbf{A}$ to weigh $\mathbf{F}_{V}$ to get updated contextually rich query vectors. Following Transformer~\cite{vaswani2017attention}, we use fully connected layers with residual connections in this process.   
\begin{flalign}
    &\mathbf{f}_{C} = \sum\limits_{H,W}\mathbf{A}*\mathbf{F}_{V} \\ 
    &\mathbf{f}_{C}=LN(\mathbf{f}_{C}+\mathbf{f}_{GQ}) \\
    &\mathbf{f}_{C}=LN(\mathbf{f}_{C}+\mathbf{FC}_{C}(\mathbf{f}_{C}))
\end{flalign}
Here, $LN$ means layer norm. It is used to stabilize the operations and prevent overfitting during training. Also, $\mathbf{*}$ represents the Hadamard product between $\mathbf{A}$ and each channel of $\mathbf{F}_{V}$. The resultant weighted $\mathbf{F}_{V}$ was averaged over the spatial dimensions. $\mathbf{f}_{C}$ is the contextually enriched feature vector that will be used in the Inference Module.

We stack multiple TX Modules together to get better context representations in the updated queries like the original Transformer architecture~\cite{vaswani2017attention, girdhar2019video}.

\subsection{Inference Module} \label{fuse}
According to~\cite{yosinski2014transferable}, fusing features from different layers of a neural network increases the expressive capability of the features. Therefore, we fuse features from our Baseline Module with the TX Module. Moreover, we refine the naive visual features $\mathbf{f}_{B}$ in the same way as equ. ~\ref{eq:HADA}.
\begin{align}
    \mathbf{f}_{BR} = {\mathbf{f}_{B}\circ \mathbf{f}_{S}\circ \mathbf{f}_{W}} 
    \label{eq:HADA2}
\end{align}
We concatenate $\mathbf{f}_{B}$, $\mathbf{f}_{BR}$, $\mathbf{f}_{C}$ together to increase the diversity in the final feature representation. With fully connected layers we make class-wise predictions for each human-object pair over this concatenated feature. Following~\cite{li2019transferable, ulutan2020vsgnet}, we also generate an interaction proposal score, $\mathbf{b}_{I}$ using the baseline features for each human-object pair. This score represents the probability of interactions between a human-object pair irrespective of the class. 

\begin{flalign}
    &\mathbf{p}_{I} = \sigma(\mathbf{FC}_{P}(\mathbf{f}_{B}  \mathbin\Vert \mathbf{f}_{BR}  \mathbin\Vert \mathbf{f}_{C})) \\ 
    &\mathbf{b}_{I}= \sigma(\mathbf{FC}_{P_{B}}(\mathbf{f}_{B}  \mathbin\Vert \mathbf{f}_{BR})) 
\end{flalign}
Here, $\sigma$ represents a sigmoid non-linearity. For each human-object pair, we achieve our final predictions by multiplying these two individual predictions. 
\begin{align}
\mathbf{p}_{HOI}= \mathbf{p}_{I}\times \mathbf{b}_{I}
\label{final_pred}
\end{align}
$\mathbf{p}_{HOI}$ has a length equal to the number of classes in considerations. 

\subsection{Loss Function} \label{loss_function}
As HOI detection is a multi-label detection task (multiple interactions can happen with the same human-object pair), almost all prior works use binary cross entropy loss for each class to train the network. However, as pointed out by ~\cite{gao2020drg, ulutan2020vsgnet}, confusing labels, missing labels, and mislabels are common in the HOI detection datasets. To handle those scenarios we utilize Symmetric Binary Cross Entropy (SCE) from ~\cite{wang2019symmetric} instead of only using binary cross entropy. This idea is derived from symmetric KL divergence and defined by:
\begin{align}
    \mathbf{SCE} = \alpha\mathbf{CE} + \beta\mathbf{RCE} = \alpha\mathbf{H}(p,q) + \beta\mathbf{H}(q,p)
\end{align}
where, \textbf{CE} is the traditional binary cross entropy, \textbf{RCE} is reverse binary cross entropy, \textbf{H} is entropy, \textbf{p} is target probability distribution and \textbf{q} is predicted probability distribution, $\alpha$ and $\beta$ are the weight values for each type of the loss. CE is useful for achieving good convergence, but it is intolerant to noisy labels. RCE is robust to noisy labels, as it compensates the penalty put on network by CE  when the target distribution is mislabeled but the network is predicting a right distribution.For more details, please refer to ~\cite{wang2019symmetric}. We select $\alpha$ and $\beta$ as 0.5 to balance both kinds of losses.

%% file: content/5_experiments.tex
\section{Experiments}
In this section, we first describe our experimental setup and implementation details. We then evaluate GTNet's performance by comparing with previously state of the art methods. Finally, we validate our design choices via different ablation studies.

\subsection{Experimental Setup}
\paragraph{\textbf{Datasets:}}

There are two widely used publicly available datasets for the HOI detection task: V-COCO~\cite{gupta2015visual}, HICO-DET~\cite{chao2018learning}. 

V-COCO is a subset of the COCO dataset~\cite{lin2014microsoft} and contains in total 10,346 images. The training, validation, and testing splits have 2,533; 2,867; and 4,946 images. Among its 29 classes, 4 do not contain any object annotations, and one of the classes has very few samples (21 images). Following previous works, we report our model's performance in the rest of the 24 classes. HICO-DET has in total 47,776 images: 38,118 for training, and 9,648 for testing. With 117 interaction classes, HICO-DET annotates in total 600 human-object interactions. Based on the number of training samples, the dataset is split into Full (all 600 HOI categories), Rare (HOI categories with sample number less than 10), and Non-Rare categories (HOI categories with sample number greater than 10)~\cite{chao2018learning}. We evaluate GTNet's performance in these categories.

\paragraph{\textbf{Evaluation Metrics}:} \label{subsection: eval}

We follow the protocol suggested by both datasets~\cite{gupta2015visual, chao2018learning} and report our model's performance in terms of mean average precision (mAP). A prediction for a human-object pair is correct if the predicted interaction matches the ground truth and both the human and object bounding boxes have an intersection over union (IOU) score of 0.5 or higher with their respective ground truth boxes. For V-COCO, there are two protocols (Scenario 1 and Scenario 2) for reporting mAP~\cite{gupta2015visual}. When there is an interaction without any object (human only), in scenario 1 the prediction would be correct if the bounding box for the object is $[0, 0, 0, 0]$, in scenario 2 in these human only cases the bounding box for the object is not considered. For HICO-DET there are two settings: default and known. In default setting all images are considered to calculate AP for a certain HOI whereas in known setting only images that contain the particular object involved in that HOI are considered.  

\begin{table}[]
\caption{Performance comparisons in the  V-COCO~\cite{gupta2015visual} test set. Many current works do not report their models' performance in Scenario 2. The "Object Detector" column indicates on which dataset each method's object detector used for inference was trained. Ground Truth indicates that the method used ground truth bounding boxes for interaction predictions. Best results in each category are marked with \textbf{bold} and the second best results in those categories are marked with \underline{underline}.}
\begin{adjustbox}{center}
\centering
\begin{tabular}{lclcc}
\hline
Method                                   & \multicolumn{1}{l}{Object Detector}               & Feature Backbone & Scenario 1    & Scenario 2    \\ \hline
DRG~\cite{gao2020drg}                    & \multirow{14}{*}{COCO}                             & ResNet50-FPN     & 51.0          & -             \\
VSGNet~\cite{ulutan2020vsgnet}           &                                                    & ResNet - 152     & 51.8          & 57.0          \\
Wan et al.~\cite{wan2019pose}            &                                                    & ResNet50-FPN     & 52.0          & -             \\
Zhong et al.~\cite{zhong2020polysemy}    &                                                    & ResNet - 152     & 52.6          & -             \\
H. Wang et.al.~\cite{wang2020contextual} &                                                    & ResNet50-FPN     & 52.7          & -             \\
Kim et al.~\cite{kim2020detecting}       &                                                    & ResNet - 152     & 53.0          & -             \\
Liu et al.~\cite{liuamplifying}          &                                                    & ResNet-50        & 53.1          & -             \\
ConsNet~\cite{liu2020consnet}            &                                                    & ResNet-50        & 53.2          & -             \\
IDN~\cite{IDN}                           &                                                    & ResNet-50        & 53.3          & \underline{60.3}          \\
OSGNet~\cite{zhang2020improved}          &                                                    & ResNet-152       & 53.4          & -             \\
Sun et al.~\cite{sun2020human}           &                                                    & ResNet - 101     & 55.2          & -             \\ 
GTNet (Ours)                                    &                                                    & ResNet-50        & \underline{56.2}          & 60.1          \\
GTNet (Ours)                           &                                                    & ResNet-152       & \textbf{58.29} & \textbf{61.85} \\ \hline
VSGNet~\cite{ulutan2020vsgnet}           & \multicolumn{1}{l}{\multirow{3}{*}{Ground Truth}} & ResNet-152       & 67.4          & 69.9          \\ 
GTNet (Ours)                                   & \multicolumn{1}{l}{}                              & ResNet-50        & \underline{71.45}          & \underline{73.50}          \\
GTNet (Ours)                           & \multicolumn{1}{l}{}                              & ResNet-152       & \textbf{73.31} & \textbf{75.14} \\ \hline
\end{tabular}
\end{adjustbox}

\label{tab:v-coco}
\end{table}
\subsection{Implementation Details}
As a backbone of GTNet, we experimented with different architectures and selected resnet-152~\cite{he2016deep} based on performance in the training set of V-COCO and HICO-DET. We also report our performance using resnet-50 \cite{he2016deep} for a fair comparison with existing methods. Output from the fourth residual block of resnet with a dimension of $1024\times 25\times 25$ was used as the the input feature map. By two separate $1\times 1$ convolutions we generate key and value with a dimension of $512\times 25\times 25$. 

For training, following the policy of previous works~\cite{gao2020drg,ulutan2020vsgnet, liu2020consnet} we use human-object bounding boxes from a pre-trained Faster-RCNN~\cite{ren2015faster} based object detector, the same as VSGNet ~\cite{ulutan2020vsgnet}. For selecting the human and object bounding boxes we use the same threshold (0.6 for human and 0.3 for object) as VSGNet. 
All projected feature vectors (baseline feature vector $\mathbf{f}_{B}$, query  vector $\mathbf{f}_{Q}$, spatial guidance vector $\mathbf{f}_{S}$, semantic guidance vector $\mathbf{f}_{W}$, guided query vector $\mathbf{f}_{GQ}$, and context rich query vector $\mathbf{f}_{C}$) have a length of 512. 
 %

Hyper parameters of the network were selected by validating on V-COCO's validation set and a small split from the training set of HICO-DET.  Our initial learning rate was 0.001 and the optimizer was Stochastic Gradient Descent (SGD) with a weight decay and a momentum of 0.9 and 0.0001. Our network was trained on GeForce RTX NVIDIA GPUs (one 2080 Ti for V-COCO, four 2080 Ti and four 1080 Ti for HICO-DET dataset) with a batch size of 8 per GPU. For each input image during training we randomly apply two augmentations from a set of augmentations (affine transformations, rotation, random cropping, random flipping, additive Gaussian noise, etc.). Our model has $\sim60M$ parameters. 

During inference, for each human-object pair, we multiply class-wise predictions $\mathbf{p}_{HOI}$ from equ.~\ref{final_pred} with the detection confidence scores of the humans and objects. Also, we apply Low grade Instance Suppressive Function (LIS)~\cite{li2019transferable} to improve the quality of the object detection confidence scores. 

%% file: content/6_results.tex
\begin{table*}[t]
\caption{Performance comparisons in the HICO-DET~\cite{chao2018learning} test set. Def and ko mean default and known settings respectively. The "Object Detector" column indicates on which dataset each method's object detector used for inference was trained. Ground Truth indicates that the method used ground truth bounding boxes for interaction predictions. Best results in each category are marked with \textbf{bold} and the second best results in those categories are marked with \underline{underline}.}
\begin{adjustbox}{width=\columnwidth,center}
\centering
\begin{tabular}{lclcccccc}
\hline
Method                                   & \multicolumn{1}{l}{Object Detector}               & Feature Backbone & Full(def)      & Rare(def)      & None-Rare(def) & Full(ko)       & Rare(ko)       & None-Rare(ko)  \\ \hline
UnionDet~\cite{kim2020detecting}         & \multirow{10}{*}{COCO+HICO-DET}                         & ResNet50-FPN     & 17.58          & 11.72          & 19.33          & 19.76          & 14.68          & 21.27          \\
IPNet~\cite{wang2020learning}            &                                                    & Hourglass-104    & 19.56          & 12.79          & 21.58          & 22.05          & 15.77          & 23.92          \\
PPDM~\cite{liao2020ppdm}                 &                                                    & Hourglass-104    & 21.1           & 14.46          & 23.09          & -              & -              & -              \\
Bansal et al.~\cite{bansal2020detecting} &                                                    & ResNet-101       & 21.96          & 16.43          & 23.62          & -              & -              & -              \\
Hou et al.~\cite{hou2020visual}          &                                                    & ResNet-50        & 23.63          & 17.21          & 25.55          & 25.98          & 19.12          & 28.03          \\
ConsNet~\cite{liu2020consnet}            &                                                    & ResNet-50        & 24.39          & 17.1           & 26.56          & -              & -              & -              \\
DRG~\cite{gao2020drg}                    &                                                    & ResNet50-FPN     & 24.53          & 19.47          & 26.04          & 27.98          & 23.11          & 29.43          \\
IDN~\cite{IDN}                           &                                                    & ResNet-50        & 26.29          & 22.61          & 27.39          & 28.24          & 24.47 & 29.37          \\ 
VSGNet~\cite{ulutan2020vsgnet}                           &                                                    & ResNet-152        & 26.54          & 21.26          & 28.12          & -          & - & -          \\ 
ATL~\cite{hou2021affordance}                                                            &                               & ResNet-50        & 27.68                          & 20.31                           & 29.89                           & 30.05                          & 22.40                           & 32.34                          \\
FCL~\cite{hou2021detecting}                                                            &                               & ResNet-50        & \underline{29.12}                          & \textbf{23.67}                           & \underline{30.75}                           & \underline{31.31}                          & \textbf{25.62}                           & \underline{33.02}                          \\ 
GTNet (Ours)                                   &                                                    & ResNet-50        & 26.78          & 21.02          & 28.50          & 28.80          & 22.19          & 30.77          \\
GTNet (Ours)                                    &                                                    & ResNet-152       & \textbf{29.71} & \underline{23.23} & \textbf{31.64} & \textbf{31.64} & \underline{24.42}          & \textbf{33.81} \\ \hline
iCAN~\cite{gao2018ican}                  & \multicolumn{1}{l}{\multirow{6}{*}{Ground Truth}} & ResNet-50        & 33.38          & 21.43          & 36.95          & -              & -              & -              \\
Li et al.~\cite{li2019transferable}      & \multicolumn{1}{l}{}                              & ResNet-50        & 34.26          & 22.9           & 37.65          & -              & -              & -              \\
Peyre et al.~\cite{peyre2019detecting}   & \multicolumn{1}{l}{}                              & ResNet-50-FPN    & 34.35          & 27.57          & 36.38          & -              & -              & -              \\
VSGNet~\cite{ulutan2020vsgnet}   & \multicolumn{1}{l}{}                              & ResNet-152    & 43.69          & 32.68          & 46.98          & -              & -              & -              \\
IDN~\cite{IDN}                           & \multicolumn{1}{l}{}                              & ResNet-50        & 43.98          & \textbf{40.27}          & 45.09          & -              & -              & -              \\ 
GTNet (Ours)                                    & \multicolumn{1}{l}{}                              & ResNet-50        & \underline{44.71}          & 33.80           & \underline{47.97}          & -          & -           & -          \\
GTNet (Ours)                                   & \multicolumn{1}{l}{}                              & ResNet-152       & \textbf{49.35} & \underline{36.93}          & \textbf{53.07} & - & - & - \\ \hline
\end{tabular}
\end{adjustbox}

\label{tab:hico-det}
\end{table*}
\subsection{Results \& Comparisons} \label{section:results}

In this section, we compare GTNet's performance with the current state of the art methods. As mentioned in Section ~\ref{subsection: eval}, we follow the evaluation protocol suggested by the V-COCO~\cite{gupta2015visual} and HICO-DET~\cite{chao2018learning} datasets. Moreover, for fair comparison we only compare our method with two stage HOI detection networks. 

For testing on V-COCO, we use the object detection results from VSGNet~\cite{ulutan2020vsgnet}, which come from an object detector trained on COCO. For testing on HICO-DET, we use the object detection results provided by DRG~\cite{gao2020drg}, which is used by the state of the art models~\cite{ulutan2020vsgnet,IDN, gao2020drg}. These detections come from an object detector trained on COCO and finetuned on the training set of HICO-DET. GTNet achieves state of the art results in all datasets.
Moreover, we use ground truth object detection results to isolate GTNet's performance from the effect of the object detector. Also, as most prior works either use resnet-152~\cite{ulutan2020vsgnet, zhang2020improved, kim2020detecting, zhong2020polysemy} or resnet-50~\cite{IDN, liu2020consnet, liuamplifying,gao2020drg} as feature backbone, we report our network's performance using both of these backbones.


In Table~\ref{tab:v-coco} and Table~\ref{tab:hico-det} GTNet's performance can be seen in V-COCO~\cite{gupta2015visual} and HICO-DET~\cite{chao2018learning} datasets. Our network outperforms all other existing methods on the V-COCO dataset in both protocols on all object detectors. The current state of the art method~\cite{sun2020human} used a fusion method to fuse features from seven branches. We clearly outperform them with both of our backbones.

Moreover, GTNet outperforms all other existing methods  with the HICO-DET object detector in the difficult default settings. Our much superior performance with ground truth object bounding boxes ($\sim6$ mAP improvement over the state of the art work) in non-rare category shows the effectiveness of our architecture. Also, though Transformer based network's needs good amount of data to train~\cite{popel2018training}, we manage to achieve second best performance in the RARE category. Actually, HICO-DET's rare category has 138 HOI classes with an average of 3 samples per class while 45 classes in rare category has only 1 sample in the training set. We expect to achieve much better performance in this category with a healthy number of samples.  
\begin{table}[b]
\caption{Ablation studies of GTNet in V-COCO test set. GTNet achieves state of the art performance with the guidance mechanism. It is interesting to observe that, in the same dataset, semantic guidance performs better than spatial guidance when tested independently. Also, it shows the effectiveness of symmetric cross entropy, interaction proposal score and data augmentation.}
\begin{adjustbox}{center}
\begin{tabular}{lcc}
\hline
                                     & Scenario 1 & Scenario 2 \\ \hline
GTNet                                & \textbf{58.29}       & \textbf{61.85}       \\ 
without Spatial Guidance             & 57.27       & 61.39       \\
without Semantic Guidance            & 53.46      & 57.10       \\
without Guidance Module              & 52.45       & 56.61       \\
without TX Module                    & 51.65      & 55.81       \\ 
without reverse cross entropy loss & 56.35       & 59.85       \\
without interaction proposal score   & 57.27      & 60.97      \\
without data augmentation            & 56.00       & 57.18      \\ 
guided by concatenation           & 57.02       & 61.20            \\ \hline
\end{tabular}
\end{adjustbox}

\label{tab:ablation_study}
\end{table}
\begin{figure*}[t]
\begin{center}
\includegraphics[width=1.0\linewidth]{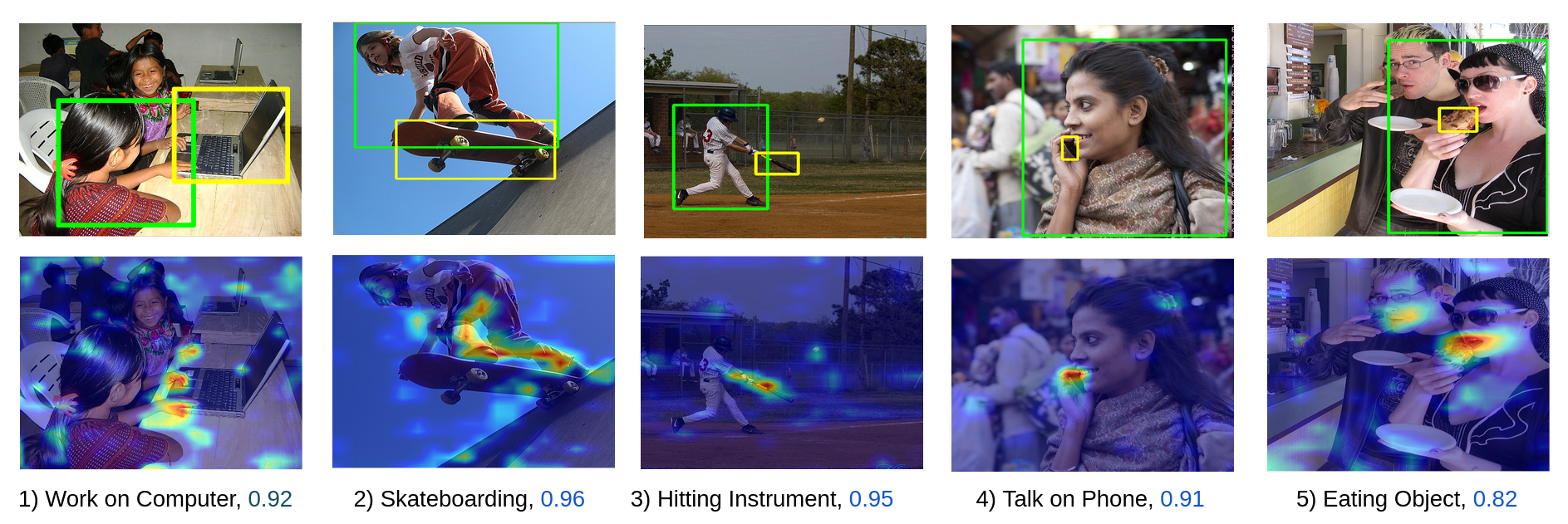}
\end{center}
\vspace{-0.2cm}
   \caption{Qualitative Results on V-COCO test set. The top row is showing images with a particular human-object pair. The bottom row is showing class activation maps~\cite{zhou2016learning} generated for these pairs along with the interaction probability for particular interactions.  The red region in the class activation map means the network is putting more attention in these areas. When there are same actions done by different human-object pairs (rightmost and leftmost images), the feature map gets activated in all relevant regions for the particular interaction. However, our pair wise guided attention strategy forces the network to consider the region significant to a particular pair.     
   }
\label{fig:qual_results}
\end{figure*} 

The closest work to our network is DRG~\cite{gao2020drg}, which utilizes a disjoint self-attention mechanism for each human and object in a dual relation graph network. By contrast, GTNet leverages the query, key, and value concept to encode pairwise contextual information in the queries along with a guiding mechanism and achieves significant improvement (7 mAP improvement in V-COCO and 5 mAP improvement in HICO-DET).
\subsection{Ablation Studies}


\noindent{\textbf{Effect of various Components and Training Policy}}: GTNet consists of several small modules and a unique training policy including the use of symmetric cross entropy loss, data augmentation, etc. In this section, we examine their effectiveness in our overall performance in V-COCO test set (Table ~\ref{tab:ablation_study}). As shown in Table ~\ref{tab:ablation_study} without the TX module, our baseline network achieves a mAP of 51.65. The performance improves slightly with the introduction of TX module. It is expected because, without the guidance mechanism, it is difficult to encode rich spatial contextual information to the visual features. With the help of the guidance mechanism, we improve our result by more than 6 mAP, highlighting the importance of the guiding mechanism in the attention framework. Also, our experiments demonstrate that semantic guidance is more effective than spatial guidance. 
\noindent
Additionally, in Table ~\ref{tab:ablation_study} we have shown GTNet's performance without reverse binary cross entropy loss (just using binary cross entropy loss), data augmentation, and interaction proposal scores. All these actually helps our network to achieve superior performance.
\noindent
Moreover, as mentioned in Section ~\ref{guide_mech} guidance can be achieved by either concatenation or product. As can be seen in Table ~\ref{tab:ablation_study}, with guidance by concatenation we achieve 57.02 mAP compared to 58.29 mAP achieved with guidance by product. 

\begin{table}[b]
\parbox{.45\linewidth}{
\caption{Performance of GTNet with different number of channels for key and value in V-COCO test set.}

\centering

\begin{tabular}{ccc}
\hline
Number of  Channels & \multicolumn{1}{l}{Scenario 1} & \multicolumn{1}{l}{Scenario 2} \\ \hline
64                  & 57.21                           & 61.1                            \\
128                 & 57.46                           & 61.6                            \\
256                 & 57.48                           & 61.25                           \\
512                 & \textbf{58.29}                           & \textbf{61.85}                           \\ \hline
\end{tabular}

\label{tab:num_of_channels}
}
 \hfill 
\parbox{.45\linewidth}{

\caption{Performance of GTNet with different size of binary spatial map, $s$ in V-COCO test set.}
\centering
\begin{tabular}{ccc}
\hline
$s$ & \multicolumn{1}{l}{Scenario 1} & \multicolumn{1}{l}{Scenario 2} \\ \hline
4                  & 56.84                           & 60.63                            \\
16                 & 57.30                           & 60.99                            \\
32                 & 57.95                           & 60.63                           \\
64                 & \textbf{58.29}                           & \textbf{61.85}                           \\ \hline
\end{tabular}
\label{tab:new_bin_map}

}
\end{table}

\noindent{\textbf{Number of Channels and Size of the Spatial Map }}: We take $1 \times 1$ convolution over the input feature map to generate key and value with different number of channels. As can be seen in Table~\ref{tab:num_of_channels} with the increasing number of channels the network seems to perform a little better. 

Similar trend can be seen in Table ~\ref{tab:new_bin_map} with the increasing size of binary spatial map, $s$ in the spatial guidance mechanism. We choose 512 as the channel size for key and value and 64 as the size of the binary spatial map considering memory constraint in our GPU.

\noindent{\textbf{Qualitative Results}}: Figure~\ref{fig:qual_results} shows some qualitative results along with the class activation maps~\cite{zhou2016learning} for particular human-object pairs. As can be seen in Figure~\ref{fig:qual_results}, GTNet correctly identifies the interactions with high confidence even when multiple interactions are happening together (image 1, 5). Moreover, from the class activation maps it is clear that our network is finding relevant context for the interacting human-object pair.   


%% file: content/7_conclusions.tex
\section{Discussion}
\subsection{Comparison with Other Attention Mechanisms}
We compare GTNet with other recent attention mechanism based HOI detection networks:  DRG~\cite{gao2020drg}, and VSGNet~\cite{ulutan2020vsgnet}. DRG~\cite{gao2020drg} considers pair-wise spatial-semantic features of humans and objects as nodes in their dual relation graph. With the use of self-attention, these features are aggregated to detect HOIs. However, we argue that using spatial-semantic features in the attention mechanism without the actual visual features will not help the network to retrieve the spatial context.  In this respect, GTNet searches the whole feature map by specific queries, which are guided by spatial-semantics features to find relevant spatial context.  Our superior quantitative results clearly validate our method. 
VSGNet~\cite{ulutan2020vsgnet} is the first attempt to utilize relative spatial configurations to refine visual features. However, VSGNet does not encode pair specific contextual information in the visual features as they take an average of the overall feature map as context. We utilize VSGNet's refinement strategy with object semantics to guide our attention mechanism. 
\vspace{-0.2cm}
\subsection{Summary}
We propose a novel guided self-attention network to leverage contextual information in detecting HOIs. Our pairwise attention mechanism utilizes a Transformer-like architecture to make visual features context-aware with the help of the guidance module. To the best of our knowledge, we are the first to propose a Guidance Module to guide the Transformer like attention mechanism to detect HOIs. We demonstrate by detailed experimentation that GTNet shows superior performance in detecting HOI and achieves the state of the art results in the standard datasets.   

%% file: content/Acknowledgement.tex
\section{Acknowledgement}

This research is partially supported by the following grants: US Army Research Laboratory (ARL) under agreement number W911NF2020157; and by NSF award SI2-SSI \#1664172. The U.S. Government is authorized to reproduce and distribute reprints for Governmental purposes notwithstanding any copyright notation thereon. The views and conclusions contained herein are those of the authors and should not be interpreted as necessarily representing the official policies or endorsements, either expressed or implied, of US Army Research Laboratory (ARL) or the U.S. Government. 


%% file: content/supp.tex
\section{Supplementary materials}

\begin{figure*}[b]
    \centering
    \includegraphics[width=\linewidth,]{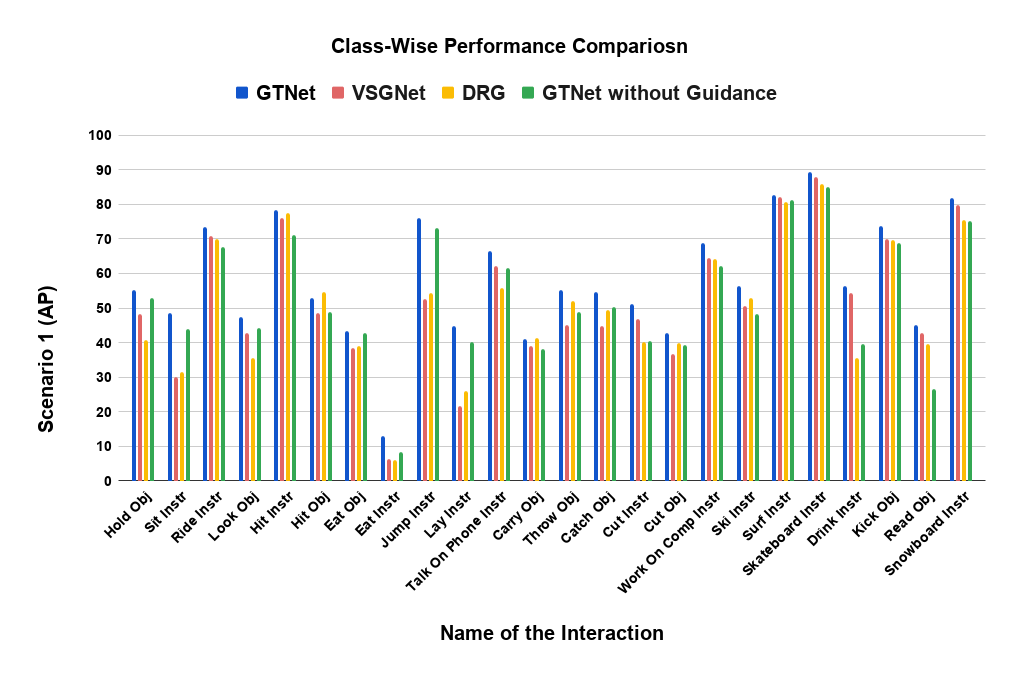}
    \vspace{-0.4cm}
    \caption{Class-wise performance comparison of GTNet with VSGNet~\cite{ulutan2020actor} and Gao et al.~\cite{gao2020drg} in V-COCO test set. We compare with the recent state of the art methods that reported class-wise AP values. Moreover, we also compare GTNet's performance without the guidance mechanism to show the effectiveness of the Guidance Module. Obj is object and instr is instrument~\cite{gupta2015visual}.}
    \label{fig:classwise}
    \end{figure*}

GTNet proposes a guided self-attention based Transformer network to detect and localize interactions between humans and objects (HOI). This supplementary material includes additional experiments to validate different design choices for the proposed GTNet and in detailed comparisons with state of the art methods along with additional qualitative results. 

\begin{table}[b]
\parbox{.45\linewidth}{
\caption{GTNet's performance on V-COCO test set for different number of heads and layers.}

\centering

\begin{tabular}{llll}
\hline
heads \textbackslash{}layers & 1    & 2     & 3    \\ \hline
1                           & 56.2 & 56.0  & 55.9 \\
2                           & 55.9 & \textbf{58.29} & 56.2 \\
4                           & 56.3 & 55.9  & 55.8 \\ \hline
\end{tabular}

\label{tab:hl}

}
 \hfill 
\parbox{.45\linewidth}{

\caption{GTNet's performance with different backbones. As can be seen, Resnet-152 achieves the best mAP in Scenario 1 of V-COCO test set.}
\hspace{0.2cm}
\centering
\begin{tabular}{lc}
\hline
Backbone                                & Scenario 1(mAP) \\ \hline
Vgg19~\cite{simonyan2014very}           & 53.4           \\
Mobile Net~\cite{sandler2018mobilenetv2} & 53.1            \\
Dense Net~\cite{iandola2014densenet}                       & 54.2            \\
Resent-34~\cite{he2016deep}             & 54.6           \\
Resnet-50~\cite{he2016deep}             & 54.6           \\
Resnet-101~\cite{he2016deep}            & 55.0           \\ 
Resnet-152~\cite{he2016deep}            & \textbf{56.4}   \\ \hline
\end{tabular}
\label{tab:backbone}

}
\end{table}




\subsection{Additional Ablation Studies}

\noindent\textbf{{Stacking of TX Modules}}: To get a better feature representation in the updated queries we stack multiple TX modules. Following~\cite{girdhar2019video}, we use the concept of heads and layers. 

\begin{figure*}[]
\begin{center}
\includegraphics[width=1\linewidth]{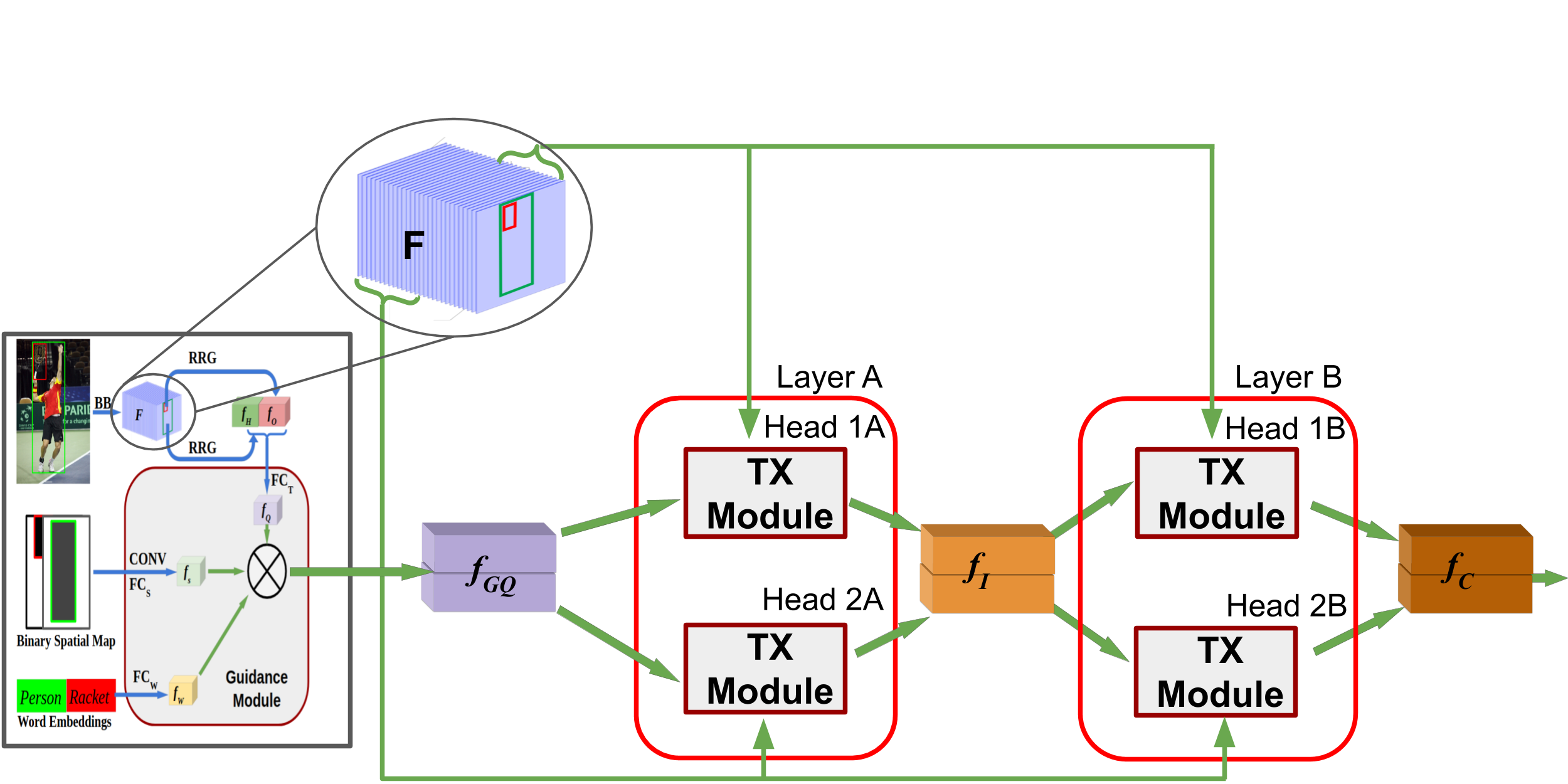}
\end{center}
\vspace{-0.4cm}
\caption{Stacking of TX Module. Two layers, two heads combination. Red colored boxes representing each layer. Each TX module inside the red box represents a head. Guided query vector $\mathbf{f}_{GQ}$ and input feature map $\mathbf{F}$ are divided into two equal parts to feed into two heads (Head 1A, Head 2A) in Layer A. The output of Layer A is concatenated to create $\mathbf{f}_{I}$ and fed to the two heads (Head 1B, Head 2B) in Layer B. The final spatial context rich feature vector is the output of Layer B ($\mathbf{f}_{C}$) which will be fed to the Inference Module. }
\vspace{-0.1cm}
\label{fig:stack}
\end{figure*}

Each layer is made up of $H$ heads. The input to the first layer is $\mathbf{f}_{GQ}$, the guided query. This input is split into $H$ parts, one for each head. Similarly, the feature map $\mathbf{F}$ is split into H parts such that the input to each head is one part of $\mathbf{F}$ and one part of $\mathbf{f}_{GQ}$. The input to each intermediate layer is the concatenated output of the previous layer and the feature map $\mathbf{F}$. All layers get the same $\mathbf{F}$ but a modified query vector. We always use same number of heads in all the layers. Fig~\ref{fig:stack} illustrates an example of a configuration with two heads and two layers. We experiment with different combinations and find that two heads and two layers combination perform best in V-COCO test set (Table~\ref{tab:hl}). Similarly, for HICO-DET we empirically choose a combination of two heads and three layers. As can be seen, the network's performance is not very sensitive to different combinations.


\noindent\textbf{{Different Backbone Networks}}:
We test with different Convolutional Neural Networks(CNN) as our feature extractor backbone networks. Table~\ref{tab:backbone} shows the performance of different backbones. We use Resnet-152 as our backbone as it achieves best result among different backbones.

\subsection{Performance Comparisons}
\noindent\textbf{Class-wise Performance}: We compare GTNet's class-wise performance in V-COCO~\cite{gupta2015visual} test set with VSGNet~\cite{ulutan2020vsgnet} and Gao et al.~\cite{gao2020drg} in Figure \ref{fig:classwise}. We also compare our network's performance without the guidance mechanism to demonstrate the effectiveness of the Guidance Module. Without the guidance mechanism, GTNet's performance is close to the state of the art methods, and with its guidance mechanism it exceeds the recent methods in most of the classes. For a few classes (jump instr, lay instr) the improvement from the recent state of the art methods are more than $10$ AP. Moreover, in a small number of classes, the object detectors perform badly. As a result the overall AP values are low in those classes. Eating utensils (instruments) are usually very small and not clearly visible in many images, so the object detectors miss them. Even with poor object detection results in eat instrument class, GTNet improves that class's performance by $\sim6$ AP over recent state of the art methods. 

\subsection{{Visualized Detections}}
\begin{figure*}[]
\begin{center}
\includegraphics[width=1\linewidth]{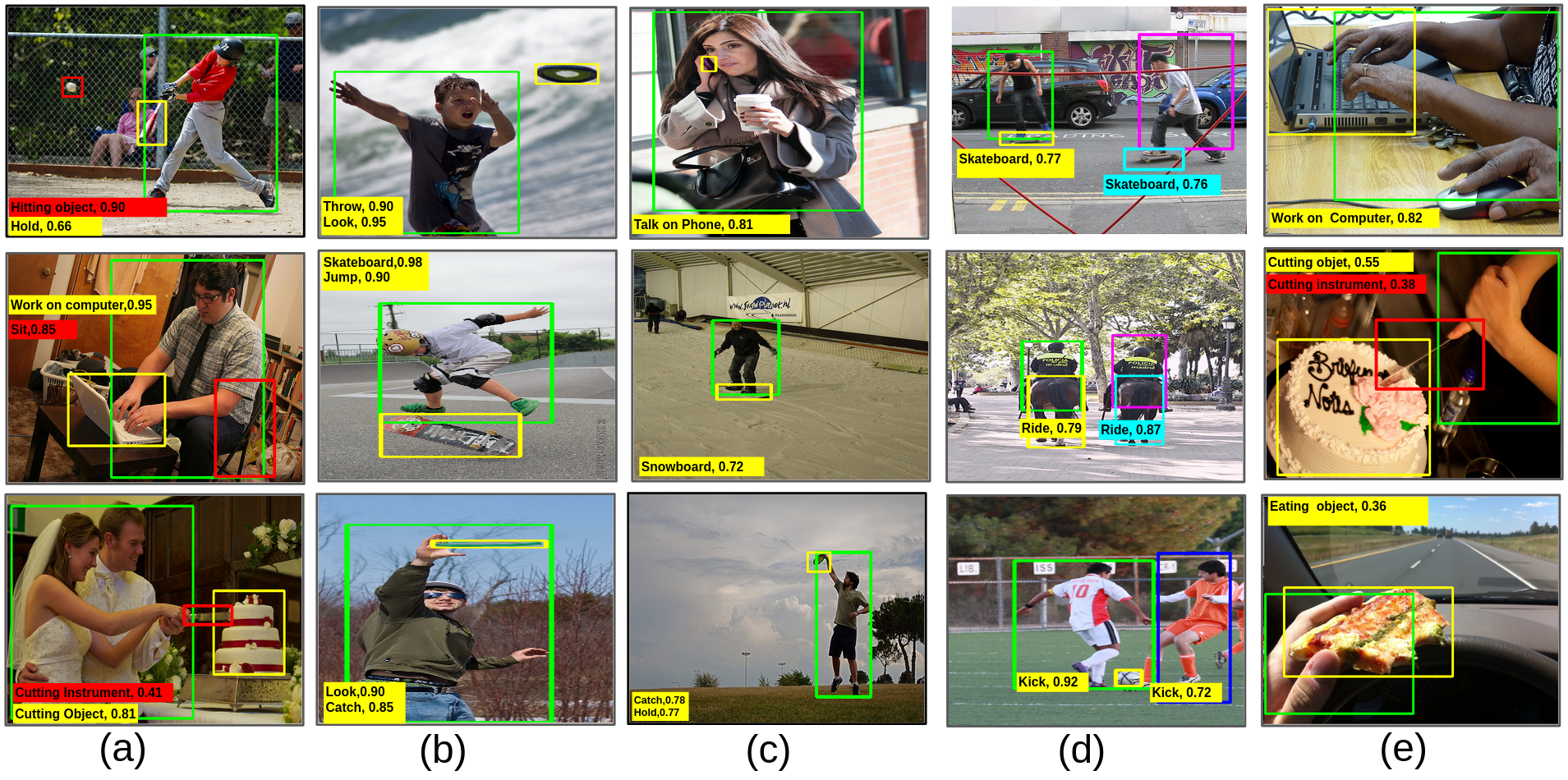}
\end{center}
\vspace{-0.25cm}
\caption{Visualized HOI detection results. Each human bounding box includes predicted interaction labels with confidence scores for that human. Interaction labels and the bounding boxes of the interacting objects are in the same color. Each column represents a different situation: (a) a single human is interacting with multiple objects, (b) no contact between interacting human and object, (c) interacting object is small or not fully visible, (d) multiple humans are interacting with either same object or different objects, (e) interacting human is not fully present in the image. }
\vspace{0cm}
\label{fig:qual_res}
\end{figure*}

In Fig.~\ref{fig:qual_res} we visualize a few predicted HOIs by GTNet. Each column in the figure represents a different situation. GTNet performs well across various situations. Moreover, as mentioned in the main paper, our network performs poorly when interacting human is not fully present in the image. This case is shown in column (e) in Fig.~\ref{fig:qual_res}.   

